\documentclass[10pt,twocolumn,letterpaper]{article}

\usepackage[pagenumbers]{cvpr} 

\usepackage[table,xcdraw]{xcolor}
\usepackage{pifont}
\usepackage{tabularray}
\usepackage{tcolorbox}
\usepackage{booktabs}
\usepackage{multirow}

\definecolor{cvprblue}{rgb}{0.21,0.49,0.74}
\usepackage[pagebackref,breaklinks,colorlinks,allcolors=cvprblue]{hyperref}
\usepackage[accsupp]{axessibility}

\newcommand{\cmark}{\ding{51}}
\newcommand{\xmark}{\ding{55}}
\def\paperID{*****} 
\def\confName{CVPR}
\def\confYear{2026}

\newcommand{\para}[1]{\vspace{.05in}\noindent\textbf{#1}\quad}

\title{VLM-Loc: Localization in Point Cloud Maps via Vision-Language Models}

\author{%
Shuhao Kang\textsuperscript{1} \quad
Youqi Liao\textsuperscript{2} \quad
Peijie Wang\textsuperscript{3} \quad
Wenlong Liao\textsuperscript{4} \quad 
Qilin Zhang\textsuperscript{5,6} \quad \\
Benjamin Busam\textsuperscript{5,6} \quad
Xieyuanli Chen\textsuperscript{7\textdagger} \quad
Yun Liu\textsuperscript{1,8,9\textdagger}\\
\textsuperscript{1}VCIP, CS, Nankai University\quad
\textsuperscript{2}Wuhan University\quad
\textsuperscript{3}CASIA\quad
\textsuperscript{4}COWAROBOT\quad \\
\textsuperscript{5}TUM\quad
\textsuperscript{6}MCML\quad
\textsuperscript{7}NUDT\quad
\textsuperscript{8}AAIS, Nankai University\quad
\textsuperscript{9}NKIARI, Shenzhen Futian\quad
}

\begin{document}
\maketitle
{
\renewcommand{\thefootnote}{}
\footnotetext[0]{\textdagger\ corresponding authors.}
}
\begin{abstract}
Text-to-point-cloud (T2P) localization aims to infer precise spatial positions within 3D point cloud maps from natural language descriptions, reflecting how humans perceive and communicate spatial layouts through language. However, existing methods largely rely on shallow text-point cloud correspondence without effective spatial reasoning, limiting their accuracy in complex environments. To address this limitation, we propose VLM-Loc, a framework that leverages the spatial reasoning capability of large vision-language models (VLMs) for T2P localization. Specifically, we transform point clouds into bird’s-eye-view (BEV) images and scene graphs that jointly encode geometric and semantic context, providing structured inputs for the VLM to learn cross-modal representations bridging linguistic and spatial semantics. On top of these representations, we introduce a partial node assignment mechanism that explicitly associates textual cues with scene graph nodes, enabling interpretable spatial reasoning for accurate localization. To facilitate systematic evaluation across diverse scenes, we present CityLoc, a benchmark built from multi-source point clouds for fine-grained T2P localization. Experiments on CityLoc demonstrate VLM-Loc achieves superior accuracy and robustness compared to state-of-the-art methods. Our code, model, and dataset are available at \href{https://github.com/MCG-NKU/nku-3d-vision}{repository}.
\end{abstract}
    
\section{Introduction}
\label{sec:intro}

\begin{figure}[t]
    \centering
    \includegraphics[width=\linewidth]{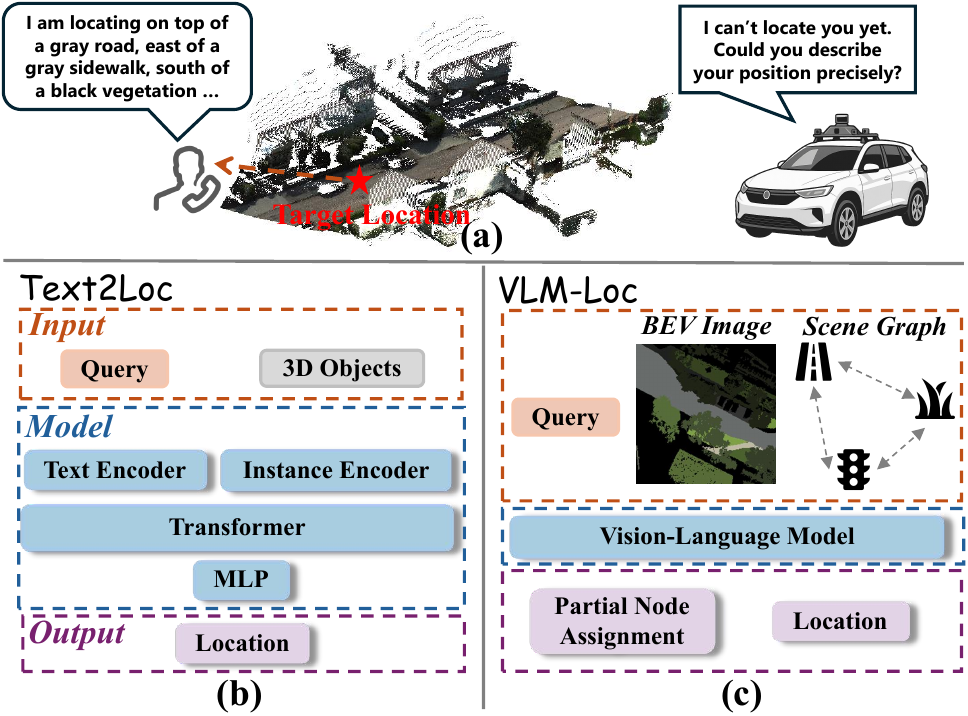}
    \vspace{-1.5em}
    \caption{
(a) illustrates the human-like logic behind text-to-point cloud localization, where spatial descriptions are used to infer the target position. 
(b) and (c) show the architectures of a typical method, Text2Loc~\cite{xia2024text2loc}, and our proposed VLM-Loc, respectively. 
}
    \label{fig:figure1}
    \vspace{-1.5em}
\end{figure}

Estimating the spatial position with natural language is a fundamental task 
in embodied intelligence~\citep{wang2024vision, intelligence2025pi_, zhou2025same, zheng2025survey} and autonomous vehicles~\citep{chen2021overlapnet,9633188,li2024diffloc, hu2025survey}. In real-world scenarios such as autonomous robotaxi services, vehicles typically rely on the Global Navigation Satellite System (GNSS) for approximate passenger localization. However, GNSS localization often suffers from degraded accuracy in urban environments due to multipath effect and atmospheric delay~\citep{teunissen2017springer}, making it difficult to identify the precise pickup spot. In such scenarios, passengers can naturally describe their surroundings using language, providing additional spatial cues for localization without relying on any visual sensors. Since point cloud maps offer detailed geometric representations of urban scenes, they are well-suited for aligning these textual cues with the physical environment. This motivates the task of text-to-point-cloud (T2P) localization, which bridges language understanding and 3D spatial perception, paving the way for future human-robot interactive localization systems, as shown in \cref{fig:figure1}(a).

Existing T2P localization approaches focus on localizing in city-scale point cloud maps using textual descriptions. Text2Pos~\cite{kolmet2022text2pos} introduced the KITTI360Pose benchmark and employed a coarse-to-fine strategy, retrieving candidate submaps before refining the position estimation. Following this paradigm, subsequent works~\cite{wang2023text,xia2024text2loc,liu2025text,xu2025cmmloc} introduced more effective model designs to improve T2P localization. However, these approaches still face several limitations: (1) during the fine localization stage, their submaps are typically restricted to small and relatively simple regions (\eg, $30\text{m}\times30\text{m}$), where the limited spatial extent inherently simplifies matching. This assumption oversimplifies real-world conditions and fails to capture the complexity of large-scale urban scenes; and (2) despite architectural improvements, these methods adopt an end-to-end position prediction paradigm without explicit reasoning, which makes structured spatial modeling challenging and limits localization accuracy, especially in complex environments.

To better understand and address these challenges, we consider what capabilities fine-grained T2P localization requires. Humans can naturally perceive and describe spatial layouts spanning tens of meters~\citep{yang2003statistical, loomis2003visual, renner2013perception}, indicating that robust localization should operate over larger and more complex regions rather than being limited to small, simplified submaps. Furthermore, overcoming the absence of explicit reasoning requires models that can interpret spatial relations expressed in language and connect them to the environment. Vision-language models (VLMs) offer a promising foundation for this goal. Equipped with strong multimodal reasoning capabilities~\citep{cai2025holisticevaluationmultimodalllms}, they can parse complex spatial descriptions and effectively align linguistic cues with visual inputs, making them suitable for fine-grained T2P localization in complex environments.

Motivated by these insights, we propose \textbf{VLM-Loc}, a novel VLM-based framework for fine-grained localization in complex local point cloud maps. As illustrated in \cref{fig:figure1} (b) and (c), unlike prior approaches~\cite{kolmet2022text2pos,xia2024text2loc} that directly learn correspondences between textual descriptions and 3D objects, VLM-Loc leverages the inherent spatial reasoning capability of large VLMs to align linguistic cues with structured map representations, achieving more interpretable and accurate localization. The point cloud map is transformed into a \textit{bird's-eye-view (BEV)} image to provide dense and structured spatial representations, while a \textit{scene graph} is simultaneously constructed to capture higher-level semantic relations among objects. Building upon these representations, we introduce \textbf{Partial Node Assignment (PNA)} mechanism that explicitly supervises the VLM to associate textual cues with their corresponding spatial nodes, thereby guiding interpretable spatial understanding and improving position estimation. To comprehensively evaluate T2P localization, we establish \textit{CityLoc}, a new benchmark specifically designed for fine-grained T2P localization in complex and diverse environments. Experimental results on \textit{CityLoc} show that the proposed VLM-Loc achieves state-of-the-art (SOTA) performance, outperforming the previous best method, CMMLoc~\cite{xu2025cmmloc}, by 14.20\% at Recall@5m on \textit{CityLoc-K} test set, demonstrating the effectiveness of our framework for accurate localization. 

Overall, our contributions are concluded as follows:
\begin{itemize}
    \item We introduce VLM-Loc, a VLM-based framework for accurate T2P localization, together with \textit{CityLoc}, a benchmark designed to systematically evaluate performance in complex 3D scenes.
    \item We transform 3D point clouds into BEV images augmented with scene graphs, bridging the modality gap between 3D point clouds and 2D VLMs and enabling effective multi-modal alignment for accurate T2P localization.
    \item We design a partial node assignment mechanism to explicitly align textual hints with graph nodes, enhancing the spatial understanding and reasoning.
\end{itemize}

\section{Related Work}
\label{sec:relatedwork}

\subsection{Localization in Point Cloud}

Localization within point cloud maps has been extensively studied in recent years.
Existing approaches mainly adopt point clouds or images as query modalities.
Pioneering point-cloud-to-point-cloud (P2P) localization methods, such as PointNetVLAD~\citep{uy2018pointnetvlad}, combine PointNet~\citep{qi2017pointnet} with NetVLAD~\citep{arandjelovic2016netvlad} for global descriptor learning. Subsequent transformer-based approaches further improve performance by modeling long-range geometric dependencies~\citep{fan2022svt, zhang2022rank, ma2022overlaptransformer, xia2023casspr}. More recently, methods like BEVPlace++~\citep{luo2025bevplace++} and RING\#~\citep{lu2025ring} enhance robustness through rotation-equivariant architectures in BEV representations.
Image-to-point-cloud (I2P) localization methods instead align image features with 3D representations through shared embeddings~\citep{cattaneo2020global, li2024vxp} or implicit reconstruction~\citep{knights2025solvr}.
In contrast, our approach employs natural language as the query modality, enabling intuitive and interpretable localization without relying on additional sensor observations, facilitating flexible interaction in real-world applications.

\subsection{3D Vision and Language}
Recent years have seen increasing interest in grounding natural language in 3D point cloud scenes.
Early works~\citep{feng2021free, jain2022bottom, ariff2026evaluating} explored text-guided 3D detection and segmentation, aiming to associate linguistic expressions with corresponding geometric structures.
Following this line, a series of studies~\citep{achlioptas2020referit3d, chen2020scanrefer, roh2022languagerefer, an2025generalized} advanced 3D language grounding through contrastive learning and transformer-based cross-modal reasoning.
These methods mainly focus on object-level grounding, \eg, identifying the referred objects or regions within a local 3D scene, thus forming the foundation for higher-level spatial understanding tasks.

Extending beyond language-driven 3D perception tasks, T2P localization requires holistic scene understanding of spatial layouts and relationships to infer specific target positions from textual descriptions. This task, first introduced by Text2Pos~\citep{kolmet2022text2pos}, aims to predict a 2-degree-of-freedom (DoF) location corresponding to the described scene context. Text2Pos formulates the problem as cell-level retrieval followed by position regression, using the KITTI360Pose dataset for evaluation. Subsequent works improved this paradigm by enhancing cross-modal representations: RET~\citep{wang2023text} and Text2Loc~\citep{xia2024text2loc} leveraged transformer-based~\citep{vaswani2017attention, liu2024vision, sun2024rethinking} encoders for text and point cloud alignment, MNCL~\citep{liu2025text} employed multi-level contrastive learning to improve boundary perception, and CMMLoc~\citep{xu2025cmmloc} modeled 3D objects using Cauchy Mixture Model priors with integrated cardinal direction cues for fine localization. 
Nevertheless, existing approaches operate on small and relatively simple submaps whose limited spatial extent simplifies localization, and they rely on direct feature matching without explicit spatial reasoning, limiting their performance in complex environments.

\subsection{Vision-Language Models}
VLMs are designed to capture holistic visual and spatial relationships across objects, scene layouts, and abstract semantic concepts.
Early works such as CLIP~\citep{radford2021learning} and ALIGN~\citep{li2021align} pioneered contrastive learning between image-text pairs, enabling robust zero-shot recognition and retrieval.
Subsequently, VLMs evolved beyond static contrastive pre-training toward richer multimodal reasoning and grounded visual understanding.
Models like BLIP-2~\citep{li2023blip} and InstructBLIP~\citep{dai2023instructblip} bridge frozen visual encoders with large language models via lightweight query transformers, achieving efficient multimodal alignment and instruction following.
More recent systems, including LLaVA series~\citep{liu2023llava, liu2024improved} and the Qwen-VL family~\citep{bai2023qwen, wang2024qwen2, bai2025qwen2, bai2025qwen3}, further enhance spatial reasoning and dense region grounding through large-scale instruction tuning and unified multimodal architectures.
Despite these advances, most existing efforts focus on perceiving and grounding real, physically present objects rather than estimating a virtual location described through language.
Motivated by this gap, our work investigates how to utilize large VLMs for text-guided localization in 3D environments, leveraging their inherent spatial reasoning abilities for accurate localization.
\begin{figure*}[t]
  \centering
  \includegraphics[width=\textwidth]{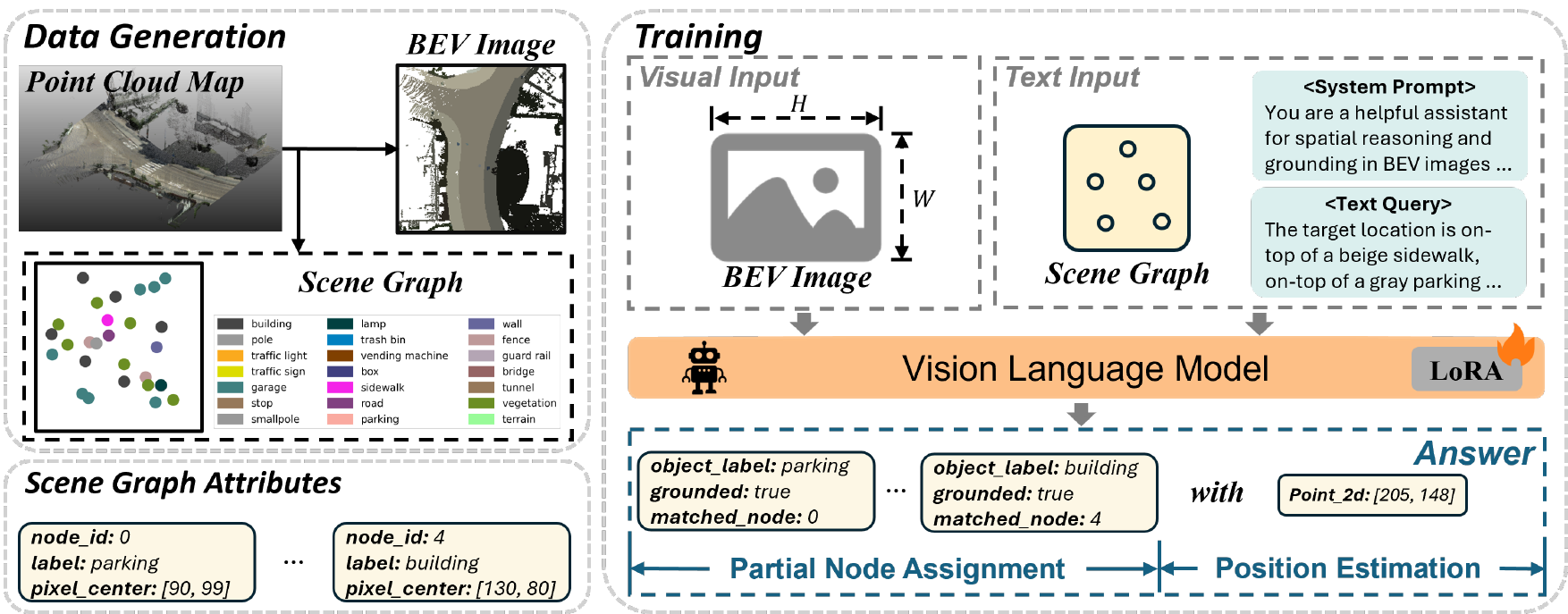}
  \caption{Overview of VLM-Loc. In the data generation stage, the point cloud map is converted into a BEV image and a scene graph, where each node encodes semantic and spatial information. During training, the BEV image is used as the visual input, and the text input includes the scene graph, system prompt, and text query. These inputs are fed into a VLM for fine-tuning, enabling it to perform partial node assignment and position estimation in an autoregressive manner. 
}
  \label{fig:overview}
  \vspace{-1em}
\end{figure*}

\section{The CityLoc Benchmark}\label{sec:benchmark}

The T2P localization is formulated on local maps for practicality. However, existing benchmark does not reflect the scale and complexity of real-world environments. The most widely used T2P localization benchmark, KITTI360Pose, was designed for large-scale point cloud localization, yet its submaps cover only small areas and contain a limited number of simple objects. This substantially simplifies the fine localization stage by restricting the search space and reducing scene clutter, yet diverges from the scenarios in which humans typically perceive and describe much larger and more complex surroundings~\citep{loomis2003visual,yang2003statistical}.

To systematically evaluate existing frameworks and our proposed method under a more realistic setting, we introduce the \textit{CityLoc} benchmark, constructed from the LiDAR point cloud of KITTI-360~\cite{liao2022kitti} and the photogrammetric point cloud of CityRefer~\citep{miyanishi2023cityrefer}, referred to as \textit{CityLoc-K} and \textit{CityLoc-C}. \textit{CityLoc-K}, derived from vehicle-mounted LiDAR scans, is used to validate the model design and assess localization performance. In contrast, \textit{CityLoc-C} is built from unmanned aerial vehicle (UAV)–based photogrammetric point clouds~\citep{hu2022sensaturban} and is used to evaluate cross-domain generalization to unseen urban scenes with different sensing modalities and semantic distributions. Such a dual-source design enables a more comprehensive evaluation of model robustness across diverse point cloud scenes.

The \textit{CityLoc} benchmark targets fine-grained T2P localization in complex environments with broader visible ranges and more diverse spatial structures, which present significant challenges for accurate pose estimation. We detail the construction process of \textit{CityLoc-K} in this section, while \textit{CityLoc-C} follows the same pipeline but adopts a tailored sampling strategy to fit its aerial viewpoint and semantic distribution. Additional details of \textit{CityLoc} are provided in the supplementary material.

\para{Map construction.}
KITTI-360~\citep{liao2022kitti} provides city-scale point cloud scenes annotated with semantic labels, instance IDs, and color information. Along the vehicle trajectory, we perform distance-based sampling to generate point cloud submaps for localization experiments. For each sampled position, a local map $\mathcal{M}$ of size $S \times S$ m is obtained by cropping all points within the corresponding spatial window centered at the sampling location. Following KITTI-360~\citep{liao2022kitti}, all instances are divided into “stuff’’ and “object’’ categories. 
For “stuff’’ categories, points within each map are grouped into discrete instances using DBSCAN~\cite{ester1996density}. 
For “object’’ categories, an instance is kept only if at least one-third of its points lie inside the map, ensuring sufficient completeness and recognizability. After processing, each local map $\mathcal{M}$ contains a set of discrete object instances, each associated with semantic labels, instance identifiers, per-point coordinates, and color attributes.

\para{Text query generation.}
Around each sampled vehicle pose, we randomly sample multiple nearby positions to serve as query locations. Each query location defines a pose cell, which represents the local region visible from that position. For each pose cell, all objects within $S \times S$ meters area centered at the query location are collected, following the same procedure used in map construction. A subset of $N_t$ objects is then randomly selected to form the textual description. For each selected object, its semantic label, color, and orientation relative to the pose cell are inserted into a predefined template to generate the natural language query.

\section{Methodology}
\noindent\textbf{Problem formulation.}\quad
Given a textual description $\mathcal{T}$ of a target position $\xi$, the goal of T2P localization is to estimate the 2D coordinates $\xi = (x, y) \in \mathbb{R}^2$ within the point cloud map $\mathcal{M}$ on the ground plane. The text query $\mathcal{T} = \{ h_i \}_{i=1}^{N_t}$ consists of $N_t$ linguistic hints describing the semantics, colors, and directions of surrounding objects relative to $\xi$. Following~\citep{sarlin2023orienternet}, we assume a locally planar ground surface, which is a reasonable approximation for urban environments at this scale.

\cref{fig:overview} presents an overview of VLM-Loc. Given a point cloud map, we first convert it into two complementary representations: a BEV image and a scene graph that capture the dense geometric layout and object-level semantics of the environment (\cref{sec:input_representation}). During localization, these representations serve as map inputs, while the text query $\mathcal{T}$ is prepended with a system prompt $s$ to guide the spatial reasoning process of the VLM. To improve spatial understanding, we introduce the PNA mechanism (\cref{sec:partial_node_assignment}), which identifies valid textual hints and grounds them to corresponding nodes in the scene graph. Based on the grounded nodes, the model predicts the target position through an autoregressive decoding procedure (\cref{sec:position_estimation}). 

\subsection{BEV Rendering and Scene Graph Generation}
\label{sec:input_representation}

\noindent\textbf{BEV image rendering.}\quad
Since VLMs are mainly pretrained on RGB images, we generate a BEV image by projecting the points of $\mathcal{M}$ onto the ground plane. This allows the models to leverage their spatial reasoning capability within a 2D layout. The map $\mathcal{M}$ contains an object set $\mathcal{O} = \{o_i\}_{i=1}^{N_o}$, where $N_o$ denotes the number of objects. Each object $o_i$ is represented by $N_i$ 3D points with RGB colors,
$
\mathcal{P}_i = \{(\mathbf{p}_{ij}, \mathbf{c}_{ij})\}_{j=1}^{N_i},
$
where $\mathbf{p}_{ij} \in \mathbb{R}^{3}$ is the per-point coordinates and $\mathbf{c}_{ij} \in \mathbb{R}^{3}$ denotes the corresponding color, and $l_i$ is its object-level semantic label.  
To obtain a compact appearance representation for each object, we compute the representative color of $o_i$ by averaging the RGB values of all points:
\begin{equation}\label{eq:color}
\bar{\mathbf{c}}_i = \frac{1}{N_i} \sum_{j=1}^{N_i} \mathbf{c}_{ij}.
\end{equation}

Next, the entire point cloud map is projected onto the ground plane and rasterized into a BEV image 
$ I \in \mathbb{R}^{H \times W \times 3} $,
covering a spatial range of $S \times S$ meters.
Each pixel in the BEV image is assigned the color $\bar{\mathbf{c}}_i$ of the object whose projected footprint occupies that grid cell.  
When a pixel contains both ``stuff'' and ``object'' categories, the ``object'' category is rendered with higher priority to ensure that foreground regions are preserved without being overwritten.

\para{Scene graph generation.}
The BEV image $I$ provides a top-down visual representation of the point cloud map $\mathcal{M}$. Although it captures the overall scene layout, it lacks explicit semantics and makes it difficult to model spatial relationships among objects. To enable more structured scene understanding for VLMs, we construct a scene graph $\mathcal{G} = (\mathcal{V}, \mathcal{E})$. Here, $\mathcal{V}$ denotes the set of object nodes and $\mathcal{E}$ represents the pairwise spatial relations among them. Each object $o_i$ is represented as a node $n_i \in \mathcal{V}$, defined as $n_i = (i, l_i, \mathbf{u}_i)$, where $i$ is the node index, $l_i$ is the semantic label, and $\mathbf{u}_i = (u_i, v_i)$ is the centroid pixel coordinates of the object on the BEV image. Since the BEV-projected coordinates $\mathbf{u}_i$ already encode relative spatial relations, we omit explicit edges $\mathcal{E}$ in practice for simplicity. This representation provides both semantic and geometric structure that aligns well with VLMs for cross-modal reasoning.

\para{Remark.}
Transforming the point cloud map into a BEV image provides a dense, rasterized representation of the environment that aligns naturally with off-the-shelf VLMs. In parallel, the scene graph serves as a structured representation that connects semantic labels to pixel locations and captures the relative spatial relationships among objects, thereby bridging the BEV image and textual instructions. Together, the BEV image and scene graph allow the VLMs to exploit both fine-grained geometric cues and high-level semantic relationships, facilitating precise spatial understanding and localization.

\subsection{Partial Node Assignment}
\label{sec:partial_node_assignment}

\begin{figure}[!t]
  \centering
  \includegraphics[width=0.5\columnwidth]{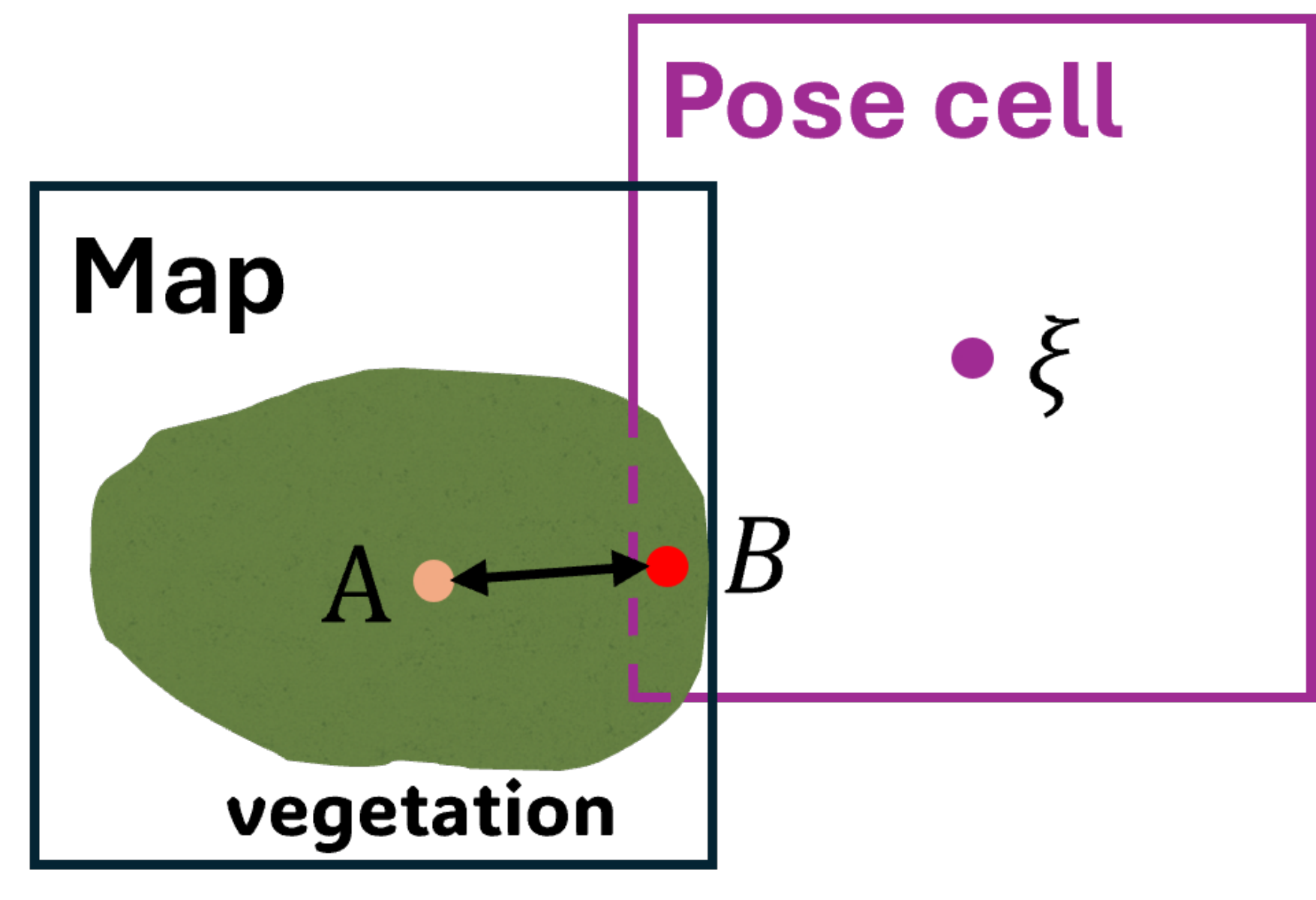}
  \vspace{-0.5em}
    \caption{Illustration of the node assignment process. PNA determines whether a textual object is groundable by comparing the distance between points A and B with the threshold $\tau$.}
  \label{fig:partial}
  \vspace{-2em}
\end{figure}

Although BEV images and scene graphs provide structured and complementary representations of the environment, T2P localization still faces two inherent challenges. First, directly regressing coordinates from these inputs does not fully leverage the structured reasoning capability of VLMs, often resulting in ambiguous spatial interpretations. Second, the map $\mathcal{M}$ covers only a limited region, some objects mentioned in the textual query may fall outside the mapped area. As a result, only part of the mentioned objects can be grounded, which leads to a \textit{partial matching} problem that further complicates accurate pose estimation.

To address this issue, we introduce the PNA mechanism, which provides supervision by explicitly aligning visible textual objects with nodes in the scene graph. For each hint $h_i$ in the textual query $\mathcal{T}$, we determine for every mentioned object whether it can be matched to a node $n \in \mathcal{V}$ in the scene graph. As illustrated in \cref{fig:partial}, consider a vegetation object whose projected center A is computed from the points of this object that fall inside the map region, and let B denote the center computed from the points of the pose cell that lie inside the region visible to the query location $\xi$. We compute the distance between A and B. If this distance is smaller than a threshold $\tau$, the textual object is considered valid, labeled as \textit{True}, and linked to the corresponding node in $\mathcal{G}$. Otherwise, it is considered invalid, labeled as \textit{False}, and the assignment for this object is set to \textit{null}.

During training, ground-truth (GT) assignments are used to supervise the model in determining whether each textual object lies inside the map $\mathcal{M}$ and in establishing textual-object correspondences. During inference, the model predicts these partial correspondences autonomously, which enables robust reasoning even when the description includes only a subset of the objects in the scene.

\para{Remark.}
PNA enables the VLM to perform interpretable localization by determining whether objects mentioned in the textual query that are visible in the map and establishing correspondences between textual query and map objects. This mechanism helps the model focus on relevant surrounding objects and understand the spatial distribution described in the text relative to the map, thereby improving localization accuracy.

\subsection{Position Estimation}
\label{sec:position_estimation}

After obtaining the node–text correspondences by the PNA module, VLM-Loc estimates the 2-DoF position $\xi$ in pixel coordinates. We incorporate position prediction into the autoregressive decoding, allowing the model to output the corresponding 2D location on BEV image in the same sequence of generation steps. This unified decoding strategy allows the model to reason consistently from correspondences to spatial coordinates. The predicted pixel location is then converted to world coordinate system.

\subsection{Loss function}
Following the autoregressive prediction paradigm of VLMs, we train VLM-Loc by maximizing the likelihood of generating the correct text–node alignments and position predictions, both of which are expressed entirely in text form. For each output token $y_t$ in the generated sequence, the training objective is the standard cross-entropy loss:
\begin{equation} \mathcal{L} = -\sum_{t=1}^{T} \log P\!\left(y_t \mid y_{<t},\, s,\, \mathcal{T},\, I,\, \mathcal{G}\right), \label{eq:loss} \end{equation}
where $T$ denotes the number of tokens in the predicted sequence.
After generating the token sequence $\{y_1, y_2, \dots, y_T\}$, the model outputs a JSON-formatted string, which we parse into structured predictions, including matched text–node pairs and the 2D pixel position.

\section{Experiments}
\label{sec:experiments}

\subsection{Experimental Setup}
\textbf{Baselines.}\quad
In experiments, we adopt representative T2P localization methods, including Text2Pos~\citep{kolmet2022text2pos}, Text2Loc~\citep{xia2024text2loc}, MNCL~\citep{liu2025text}, and CMMLoc~\citep{xu2025cmmloc} as baselines. For fair comparison, all baselines are retrained on \textit{CityLoc-K} using only the localization modules, with official implementations and identical training configurations.

\para{Evaluation metrics.} Following prior localization works~\citep{kolmet2022text2pos,sarlin2023orienternet}, we evaluate localization performance using Recall@$K$ m, which measures the percentage of samples whose predicted positions are within $K$ meters of the GT location.  Results are reported for $K \in \{5, 10, 15\}$.

\para{Implementation details.}
We employ Qwen3-VL-8B-Instruct~\cite{yang2025qwen3} as the base model for the VLM-Loc framework. Training is performed using the Swift framework~\cite{zhao2025swift} with LoRA-based parameter-efficient tuning~\cite{hu2022lora}. We insert LoRA adapters with rank $r{=}8$ and scaling factor $\alpha{=}16$ into all linear layers. The parameters of the vision encoder, vision adapters and language backbone remain frozen, and only LoRA parameters are updated during training, preserving pretrained cross-modal alignment while adapting the model to the domain gap between rendered BEV and natural images. The model is trained for 2 epochs with a global batch size of 4 on 8 NVIDIA RTX 4090 GPUs. We use the AdamW optimizer~\cite{loshchilov2017decoupled} with a learning rate of $1\times10^{-4}$ and a warm-up ratio of 0.05. All experiments are conducted in bfloat16 precision.

For each BEV image $I$, the resolution is set to $H = W = 224$, corresponding to a spatial coverage of $S = 50$. Each textual query $\mathcal{T}$ contains $N_t = 6$ hints. A dynamic threshold $\tau$ is used for different semantic categories: 5 m for ``object'' classes and 15 m for ``stuff'' classes.

\begin{table}[t]
\centering
\small
\setlength{\tabcolsep}{2.8pt}
\renewcommand{\arraystretch}{1.05}
\begin{tabular}{lccc|cc}
\toprule
\multirow{2}{*}{\#} &
\multicolumn{2}{c}{Input} &
\multicolumn{1}{c|}{Output} &
\textit{CityLoc-K Val} & \textit{CityLoc-K Test}\\
\cmidrule(lr){2-3} \cmidrule(lr){4-4} \cmidrule(lr){5-5} \cmidrule(lr){6-6}
 & BEV & SG & PNA & R@5/10/15m & R@5/10/15m \\
\midrule
(a) & \cmark & \xmark & \xmark & 13.04/32.79/52.31 & 13.21/33.86/51.40 \\
(b) & \xmark & \cmark & \xmark & 26.75/52.94/71.67 & 24.62/51.25/69.46 \\
(c) & \xmark & \cmark & \cmark & \underline{33.69}/\underline{59.48}/75.51 & \underline{32.34}/\underline{61.34}/\underline{74.94} \\
(d) & \cmark & \cmark & \xmark & 29.06/59.37/\underline{77.01} & 29.79/57.57/73.78 \\
\midrule
\rowcolor{gray!20}
(e) & \cmark & \cmark & \cmark & 
\textbf{36.23}/\textbf{63.66}/\textbf{77.77} & 
\textbf{35.91}/\textbf{63.81}/\textbf{76.79} \\
\bottomrule
\end{tabular}
\vspace{-0.5em}
\caption{
Ablation study on each component.  
Input: BEV = BEV image, SG = scene graph.  
Output: PNA = partial node assignment.  
Best results are in \textbf{bold}, and second-best results are \underline{underlined}.
}
\label{tab:ablation_compact}
\vspace{-0.5em}
\end{table}

\begin{table}[!t]
\centering
\small
\setlength{\tabcolsep}{4pt}
\renewcommand{\arraystretch}{1.15}
\begin{tabular}{l c c}
\toprule
\multirow{2}{*}{Strategy} &
\multicolumn{1}{c}{\textit{CityLoc-K Val}} &
\multicolumn{1}{c}{\textit{CityLoc-K Test}} \\
\cmidrule(lr){2-2} \cmidrule(lr){3-3}
 & R@5/10/15 m & R@5/10/15 m \\
\midrule
Full & 18.23/44.70/63.10 & 17.81/41.55/60.67 \\
\rowcolor{gray!20}
Partial & \textbf{36.23}/\textbf{63.66}/\textbf{77.77} & \textbf{35.91}/\textbf{63.81}/\textbf{76.79} \\
\bottomrule
\end{tabular}
\vspace{-0.5em}
\caption{Ablation study on partial and full node assignment.}
\label{tab:partial_full}
\vspace{-0.5em}
\end{table}

\begin{table}[!t]
\centering
\small
\setlength{\tabcolsep}{3pt}
\renewcommand{\arraystretch}{1.05}
\begin{tabular}{lccc|cc}
\toprule
\multirow{2}{*}{\#} &
\multicolumn{3}{c|}{Components} &
\textit{CityLoc-K Val} & \textit{CityLoc-K Test} \\
\cmidrule(lr){2-4} \cmidrule(lr){5-5} \cmidrule(lr){6-6}
 & Sem. & Color & Dire.  & R@5/10/15m & R@5/10/15m \\
\midrule
(a) & \cmark & \xmark & \xmark & \underline{18.74}/43.17/63.15 & 16.93/40.81/60.22 \\
(b) & \cmark & \cmark & \xmark &
18.28/\underline{43.96}/\underline{64.50} &
\underline{18.01}/\underline{42.95}/\underline{61.47} \\
\midrule
\rowcolor{gray!20}
(c) & \cmark & \cmark & \cmark &
\textbf{36.23}/\textbf{63.66}/\textbf{77.77} &
\textbf{35.91}/\textbf{63.81}/\textbf{76.79} \\
\bottomrule
\end{tabular}
\vspace{-0.5em}
\caption{
Ablation study on text query components.  
}
\label{tab:query_component}
\vspace{-1.5em}
\end{table}

\begin{table}[!t]
\centering
\small
\setlength{\tabcolsep}{3pt}
\renewcommand{\arraystretch}{1.15}
\begin{tabular}{l c c}
\toprule
\multirow{2}{*}{Model} &
\multicolumn{1}{c}{\textit{CityLoc-K Val} } &
\multicolumn{1}{c}{\textit{CityLoc-K Test}} \\
\cmidrule(lr){2-2} \cmidrule(lr){3-3}
 & R@5/10/15 m & R@5/10/15 m \\
\midrule
Qwen3-VL-2B-Instr. & 35.78/64.22/79.97 & 34.70/63.19/76.49\\
Qwen3-VL-4B-Instr. & 35.50/64.28/80.76 & 34.23/61.37/75.18 \\
Qwen3-VL-32B-Instr. & 
\textbf{39.84}/\textbf{67.72}/\textbf{82.05} & 
\textbf{41.05}/\textbf{67.47}/\textbf{79.39} \\
InternVL3.5-8B~\cite{wang2025internvl3} &
\underline{38.32}/\underline{65.41}/\underline{81.21} &
\underline{38.14}/63.66/\underline{77.25} \\
\midrule
\rowcolor{gray!20}
Qwen3-VL-8B-Instr. & 36.23/63.66/77.77 & 35.91/\underline{63.81}/76.79 \\
\bottomrule
\end{tabular}
\caption{Ablation study on the effect of different VLM backbones.}
\label{tab:vlm_comparison}
\vspace{-2em}
\end{table}

\begin{table*}[t]
\centering
\small
\setlength{\tabcolsep}{3.0pt}
\renewcommand{\arraystretch}{1.1}
\begin{tabular}{l | ccc | ccc}
\toprule
\multirow{2}{*}{Method}  & \multicolumn{3}{c|}{\textit{CityLoc-K Val}} & \multicolumn{3}{c}{\textit{CityLoc-K Test}} \\
\cmidrule(lr){2-4} \cmidrule(lr){5-7}
 & R@5\,m & R@10\,m & R@15\,m & R@5\,m & R@10\,m & R@15\,m \\
\midrule
Text2Pos~\cite{kolmet2022text2pos}       & 16.48 & 40.69 & 62.92 & 14.62 & 38.27 & 59.55 \\
Text2Loc~\cite{xia2024text2loc}          & 18.91 & 45.26 & 64.28 & 17.97 & 41.22 & 61.50 \\
MNCL~\cite{liu2025text}                  & 19.30 & 45.94 & 64.50 & 18.76 & 42.63 & 62.58 \\
CMMLoc~\cite{xu2025cmmloc}               & \underline{20.77} & \underline{48.65} & \underline{67.89} & \underline{21.71} & \underline{46.67} & \underline{66.00} \\
\midrule
\rowcolor{gray!20}
\textbf{VLM-Loc}              & \textbf{36.23} {\scriptsize \color{green!50!black}(+15.46)} & 
                                        \textbf{63.66} {\scriptsize \color{green!50!black}(+15.01)} & 
                                        \textbf{77.77} {\scriptsize \color{green!50!black}(+9.88)} & 
                                        \textbf{35.91} {\scriptsize \color{green!50!black}(+14.20)} & 
                                        \textbf{63.81} {\scriptsize \color{green!50!black}(+17.14)} & 
                                        \textbf{76.79} {\scriptsize \color{green!50!black}(+10.79)} \\
\bottomrule
\end{tabular}
\caption{
Localization results of VLM-Loc and baseline methods on \textit{CityLoc-K}. Green numbers indicate improvements over baselines. 
}
\label{tab:main_results}
\end{table*}

\subsection{Ablation Study}
\label{sec:ablation}

This section provides ablation studies to evaluate: (1) the effectiveness of each component in VLM-Loc; (2) the impact of the PNA mechanism; (3) the effect of different text query construction strategies; and (4) the effect of various VLM backbones. Ablation studies are conducted on \textit{CityLoc-K}.

\para{Effect of components.}
We conduct ablation studies to assess the contribution of each component in VLM-Loc, including the BEV image, scene graph (SG), coordinate prediction (C), and the proposed PNA mechanism, as summarized in \cref{tab:ablation_compact}.
Variant (a), which uses only the BEV image, performs poorly, showing that dense appearance cues alone cannot capture object relationships and support accurate localization.
Variant (b), which relies solely on the SG, significantly outperforms (a), indicating that relational structure is more effective for position estimation.
Introducing PNA on top of SG, as in (c), further boosts performance, improving Recall@5m by 6.94\% on the validation set and 7.72\% on the test set. This confirms that explicit node-level grounding strengthens the reliability of coordinate prediction.
Variant (d), which combines the BEV image with the scene graph, further improves performance by 2.31\% and 5.17\% at Recall@5m over the SG-only baseline. This improvement indicates that integrating dense visual cues with relational structure provides the model with more complete spatial information.
Finally, the full model (e), integrating BEV, SG, and PNA, achieves the best overall performance, demonstrating the synergy between multimodal inputs and explicit grounding for T2P localization.

\para{Comparison between partial and full node assignment.}
In our PNA strategy, an object mentioned in a hint $h$ is considered groundable when the distance between the centroid of its visible region in the map $M$ and the centroid of its visible region in the pose cell is smaller than a threshold $\tau$. Otherwise, the object is treated as not groundable. To evaluate its effectiveness, we compare PNA with a full node assignment variant. For this variant, any object mentioned in the textual query is always assigned to a node in the scene graph as long as there exists at least one node with the same semantic label. In particular, the object is forced to match the nearest node with the same label, even when their centroid distance exceeds $\tau$, rather than being left unassigned. As shown in \cref{tab:partial_full}, the proposed partial node assignment consistently outperforms the full assignment on both the validation and test sets. Specifically, it improves Recall@5m by 18.00\% and 18.10\%, respectively, demonstrating that explicitly accounting for partial visibility allows the model to focus on truly observable objects, thereby enhancing node grounding and ultimately benefiting position estimation.

\para{Effect of text query components.}
As shown in \cref{tab:query_component}, we analyze the contribution of semantic, color, and directional cues in the text queries. Removing both color and directional information (Variant (a)) leads to a severe performance drop, reducing Recall@5m from 36.23\% to 18.74\% on the validation set and from 35.91\% to 16.93\% on the test set. When keeping color but removing directional cues (Variant (b)), the performance remains significantly lower than the full setting, with Recall@5m of 18.28\% on validation and 18.01\% on test. These results show that directional cues play the dominant role in spatial reasoning, as removing them nearly halves the performance even when color information is preserved, while color provides complementary appearance grounding that further enhances localization when combined with directional cues.

\para{Effect of VLM backbones.}
We evaluate the impact of different VLM architectures and model sizes on localization performance, as summarized in \cref{tab:vlm_comparison}.
By leveraging either InternVL3.5~\cite{wang2025internvl3} or the Qwen3-VL-Instruct models~\cite{bai2025qwen3}, VLM-Loc achieves strong performance, showing that diverse VLM architectures are compatible with and effective for VLM-Loc.
Within the Qwen3-VL family, the 2B, 4B, and 8B variants achieve comparable results, while the 32B model shows a significant gain, reaching 39.84\% Recall@5m on the validation set.
This scaling trend indicates that VLM-Loc consistently benefits from the enhanced multimodal reasoning capacity of larger models.

\subsection{Results}

\begin{figure}[!t]
  \centering
  \includegraphics[width=0.9\columnwidth]{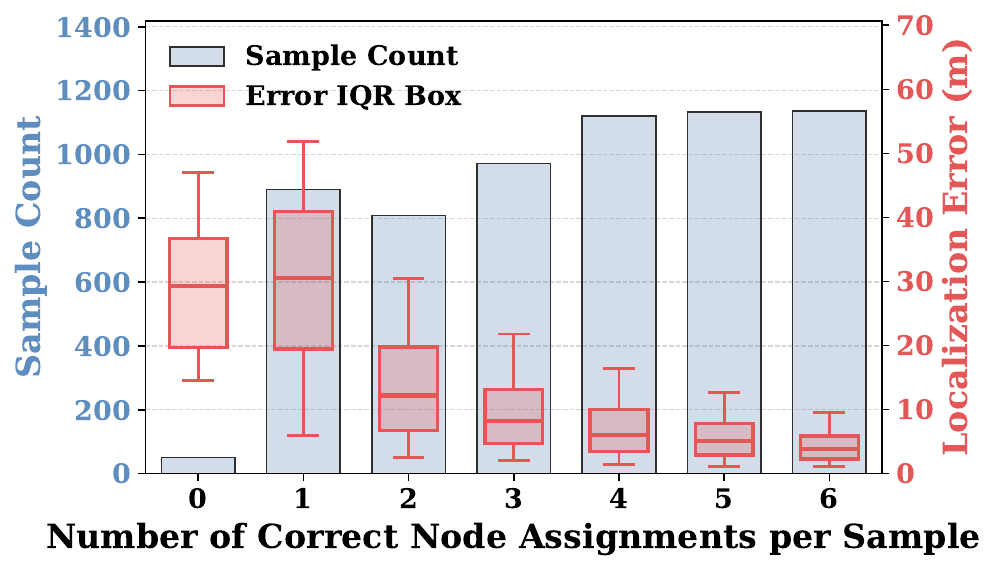}
  \vspace{-0.5em}
  \caption{Relationship between localization error and the number of correctly assigned nodes on the \textit{CityLoc-K} test set. More correct node assignments correspond to lower localization errors.}
  \label{fig:node_assign}
  \vspace{-1.5em}
\end{figure}

\begin{figure*}[!t]
  \centering
  \includegraphics[width=\textwidth]{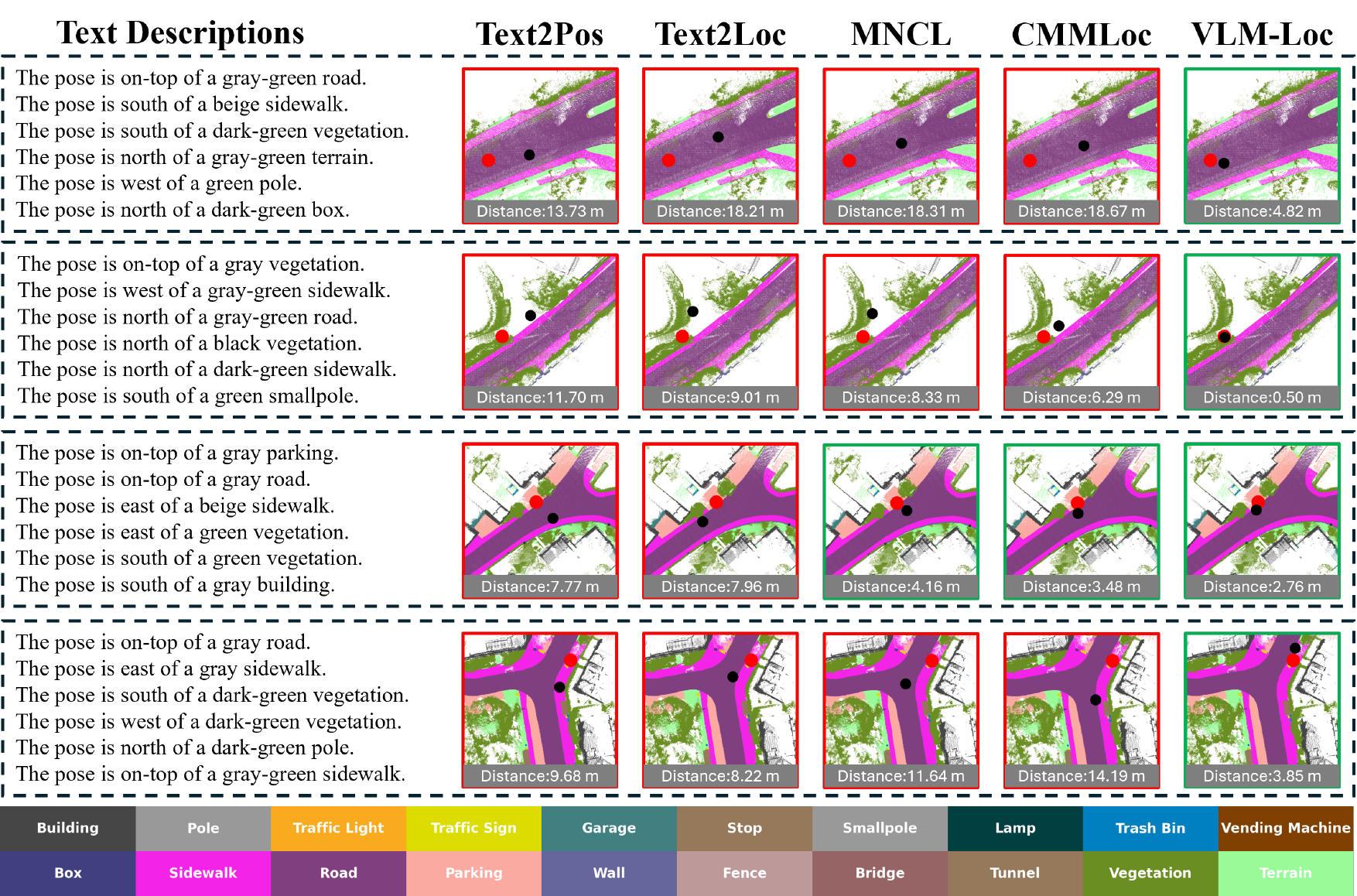}
  \caption{Qualitative results of VLM-Loc and baseline methods on the \textit{CityLoc-K}.
Each example visualizes the predicted and GT positions on colorized BEV maps rendered with semantic labels.
The red circles \textcolor{red}{\ding{108}} and black circles \textcolor{black}{\ding{108}} denote the GT and predicted positions, respectively. The localization error is shown below each image, and green/red borders indicate localization error below/above 5 m.}
  \label{fig:qualitative_result}
  \vspace{-1em}
\end{figure*}

\textbf{Node assignment results.} \quad
\cref{fig:node_assign} presents the distribution of samples according to the number of correctly assigned nodes on the \textit{CityLoc-K}. Most samples achieve three or more correct assignments, indicating that VLM-Loc is generally effective at grounding textual cues to corresponding scene graph nodes. On the right, the interquartile range (IQR)~\cite{dekking2005modern} illustrates the distribution of localization errors. A clear trend emerges: as the number of correctly assigned nodes increases, both the median error and the spread of the error distribution decrease substantially, with notably stable performance once four or more nodes are correctly grounded. This strong correlation highlights the importance of accurate node assignment, as reliable grounding of textual descriptions directly leads to precise localization.

\para{Localization accuracy.}
Localization results of the baseline methods and our proposed VLM-Loc are presented in \cref{tab:main_results}. 
On \textit{CityLoc-K}, VLM-Loc achieves the best localization performance and significantly outperforms the strongest baseline CMMLoc, with improvements of 15.46\% and 14.20\% in Recall@5m on the validation and test sets, respectively. Qualitative examples in \cref{fig:qualitative_result} further show that VLM-Loc consistently surpasses all baselines across diverse scenes. Results show that VLM-Loc attains strong spatial understanding through BEV and scene graph representations, and performs structured reasoning via the PNA mechanism, yielding more accurate and interpretable localization than existing baselines. More qualitative results are available in the supplementary material.

\begin{table}[!t]
\centering
\small
\setlength{\tabcolsep}{5pt}
\begin{tabular}{l | ccc}
\toprule
\multirow{2}{*}{Method} & \multicolumn{3}{c}{\textit{CityLoc-C}} \\
\cmidrule(lr){2-4}
 & R@5\,m & R@10\,m & R@15\,m \\
\midrule
Text2Pos~\cite{kolmet2022text2pos} & 8.11 & 27.21 & 50.01 \\
Text2Loc~\cite{xia2024text2loc}    & 9.45 & 29.44 & 51.17 \\
MNCL~\cite{liu2025text}            & \underline{13.68} & \underline{35.93} & 53.78 \\
CMMLoc~\cite{xu2025cmmloc}         & 11.68 & 34.79 & \underline{54.71} \\
\midrule
\rowcolor{gray!20}
\textbf{VLM-Loc}                   
& \textbf{21.37}
& \textbf{49.12}
& \textbf{68.26} \\
\bottomrule
\end{tabular}
\caption{
Generalization results on the \textit{CityLoc-C} benchmark.
}
\label{tab:generalization}
\vspace{-2em}
\end{table}

\para{Generalization ability.}
To evaluate cross-domain generalization, we directly transfer the models trained on \textit{CityLoc-K} to the \textit{CityLoc-C} split and assess their localization accuracy without extra fine-tuning. The point cloud data of \textit{CityLoc-C} originates from the SensatUrban~\citep{hu2022sensaturban} dataset, which contains photogrammetric point clouds captured by drones and thus exhibit characteristics markedly different from the vehicle-mounted LiDAR data used in \textit{CityLoc-K}. As shown in \cref{tab:generalization}, VLM-Loc achieves substantially higher recall across all thresholds (21.37\%, 49.12\%, and 68.26\% at 5/10/15m), significantly outperforming prior  methods. These results demonstrate that VLM-Loc generalizes effectively to unseen environments and heterogeneous point cloud sources.

\section{Conclusion}
\label{sec:conclusion}

In this paper, we present VLM-Loc, a VLM-based framework for localization in point cloud maps using text queries. By leveraging the visual and textual understanding and the strong spatial reasoning capability of VLMs, our approach enables accurate localization from language cues in complex environments. VLM-Loc converts point cloud maps into BEV images and scene graphs, and introduces a partial node assignment mechanism that explicitly aligns textual cues with spatial nodes. We also establish the \textit{CityLoc} benchmark to evaluate fine-grained T2P localization. Experiments on \textit{CityLoc} demonstrate that VLM-Loc achieves substantial improvements in localization accuracy and robustness over existing baselines.

\vspace{.3in}
\noindent\textbf{Future work.}\quad
We believe that this work paves the way for two promising research directions:
(1) Strengthening multi-step reasoning and scene understanding so that models can handle longer and more compositional textual descriptions in complex outdoor environments.
(2) Building upon the \textit{CityLoc} benchmark to move from passive localization toward an active agent~\citep{li2025theory} that unifies localization with planning and navigation in unseen environments, improving the ability to interact effectively with its surroundings.

\section*{Acknowledgements}
This work is supported in part by the Fundamental Research Funds for the Central Universities (Nankai University, No. 070-63253235) and in part by NSFC (No. 62576176).

{
    \small
    \bibliographystyle{ieeenat_fullname}
    \bibliography{main}
}

\clearpage
\setcounter{page}{1}
\maketitlesupplementary
\setcounter{section}{0}
\renewcommand\thesection{\Alph{section}}

\section{Overview}
\label{sec:overview}

In the supplementary material, we provide additional benchmark construction and statistics (\cref{sec:benchmark_details}), implementation details of textual query and system prompt (\cref{sec:implementation_details}), and further visualization results (\cref{sec:more_vis}).

\begin{figure}[!t]
  \centering
  \includegraphics[width=\columnwidth]{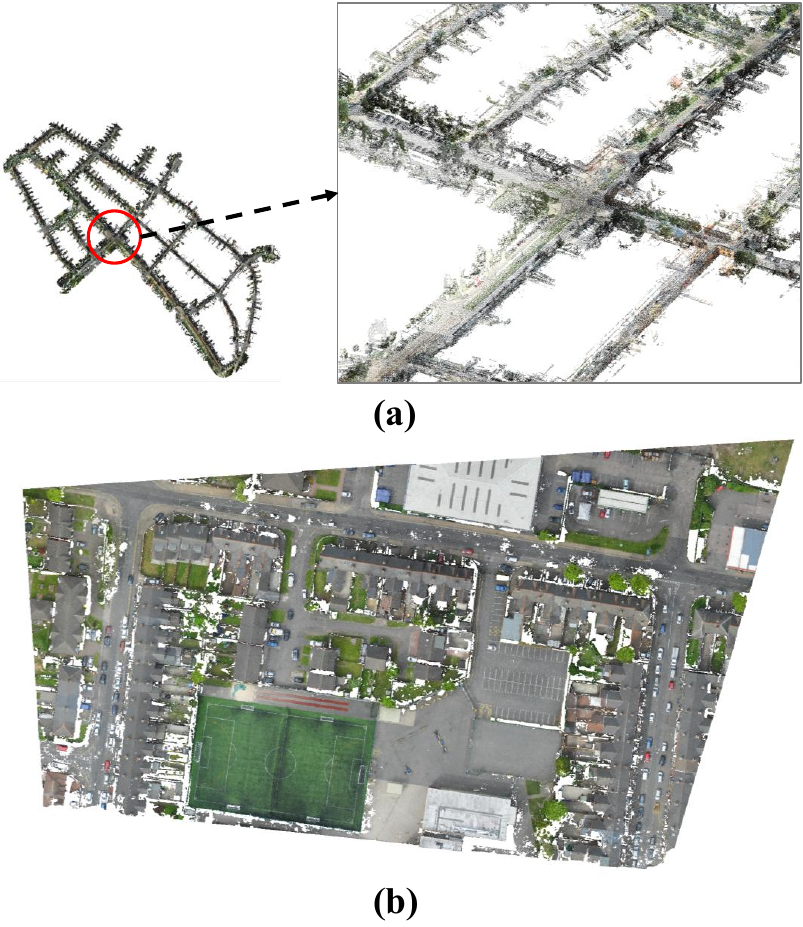}
\caption{Example point clouds from the \textit{CityLoc} benchmark. 
(a) A roadside LiDAR scene from KITTI-360~\cite{liao2022kitti}. 
(b) A photogrammetric urban block from CityRefer~\cite{miyanishi2023cityrefer}.}
  \label{fig:scene_view}
  \vspace{-1.5em}
\end{figure}

\section{The \textit{CityLoc} Benchmark Details}
\label{sec:benchmark_details}

Our \textit{CityLoc} benchmark consists of two subsets, including \textit{CityLoc-K} for training, validation and testing, \textit{CityLoc-C} for cross-domain testing, as shown in Fig.~\ref{fig:scene_view}. The two splits differ significantly in semantic composition, sensing modality, point cloud characteristics, and geographic region, offering a diverse and challenging setting for evaluating the generalization ability of T2P localization models. 

In this section, we provide additional details that complement the description in \cref{sec:benchmark}. We first describe the construction procedures of \textit{CityLoc-K} and \textit{CityLoc-C} in \cref{sec:cityloc_k} and \cref{sec:cityloc_c}, respectively. In \cref{sec:benchmark statistics} we then present the statistics of the proposed \textit{CityLoc} benchmark.
 
\subsection{CityLoc-K Construction}
\label{sec:cityloc_k}

\textbf{Data source.} \quad
\textit{CityLoc-K} is constructed based on the KITTI-360 dataset\footnote{Available under the Creative Commons \href{https://creativecommons.org/licenses/by-nc-sa/3.0/}{Attribution-NonCommercial-ShareAlike 3.0 license}.}, which contains large-scale LiDAR point clouds collected by vehicle-mounted sensors along urban roads in Karlsruhe. We use 5 training sequences (\textit{00, 02, 04, 06, 07}), 1 validation sequence (\textit{10}), and 3 testing sequences (\textit{03, 05, 09}).

\para{Map center sampling.}
When constructing the maps, we sample map centers from the vehicle trajectory. Specifically, we perform distance-based subsampling along the trajectory to ensure that any two selected centers are at least 10 m apart. This ensures that sampling remains evenly distributed across the entire trajectory, preventing points from clustering in specific regions and avoiding biased localization outcomes.

\para{Query location sampling.}
For each sampled map center, we further generate four query positions by randomly perturbing its horizontal coordinates within $\pm 15\,\text{m}$ along both the East and North directions, thereby expanding the number of query samples and positional diversity.

\subsection{CityLoc-C Construction}
\label{sec:cityloc_c}

\textbf{Data source.} \quad
\textit{CityLoc-C} is derived from SensatUrban~\cite{hu2022sensaturban} and CityRefer~\cite{miyanishi2023cityrefer}\footnote{Both datasets are released under the \href{https://mit-license.org/}{MIT License}.}. 
For convenience, we denote blocks in Birmingham and Cambridge as \textit{B\#} and \textit{C\#}, respectively. 
The dataset contains high-resolution photogrammetric point clouds with nearly three billion points collected across three UK cities, covering an area of 7.6~km\textsuperscript{2}. 
We use the validation and test splits from Birmingham and Cambridge, selecting 4 blocks in Birmingham (\textit{B0, B5, B6, B12}) and 7 blocks in Cambridge (\textit{C2, C3, C8, C10, C14, C21, C26}) for cross-city generalization analysis.

\para{Map center sampling.}
CityRefer does not provide the sensor trajectory of the data acquisition process.  Therefore, for map center sampling, we perform grid sampling on each point cloud block and retain maps that contain more than six objects.  For query location sampling, we follow the same strategy used for \textit{CityLoc-K}, as described in \cref{sec:cityloc_k}.

\subsection{Benchmark Statistics}\label{sec:benchmark statistics}

\textbf{Dataset-Level Statistics.}\quad 
\textit{CityLoc-K} contains 2,767, 300, and 1,027 maps, together with 16,113, 1,772, and 6,109 queries for training, validation, and testing, respectively. These maps span areas of 1.66, 0.19, and 0.60~km\textsuperscript{2}.
\textit{CityLoc-C} includes 875 maps and 4,487 queries, covering 0.90~km\textsuperscript{2} for cross-domain testing.
The reported map area is computed as the union of the minimum bounding boxes of all maps.

\begin{figure}[!t]
  \centering
  \includegraphics[width=\columnwidth]{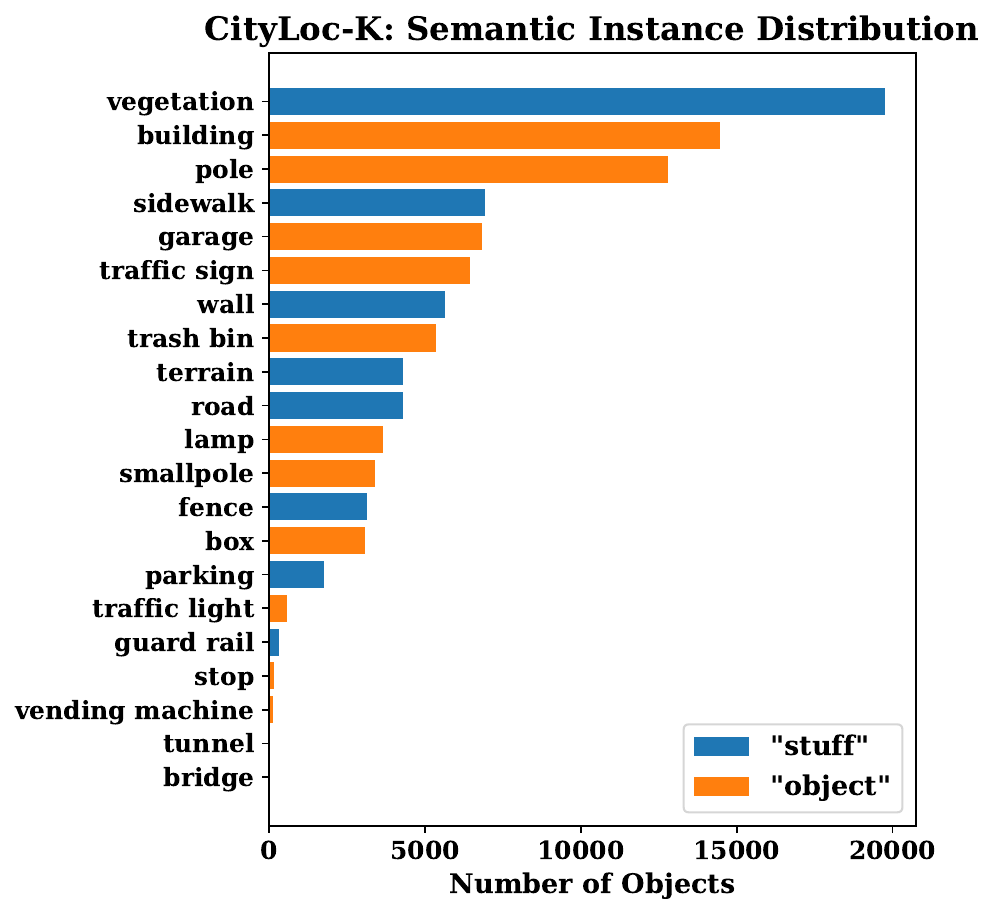}
    \caption{Semantic instance distribution in \textit{CityLoc-K}.}
  \label{fig:cityloc_k_distribution}
\end{figure}

\begin{figure}[!t]
  \centering
  \includegraphics[width=\columnwidth]{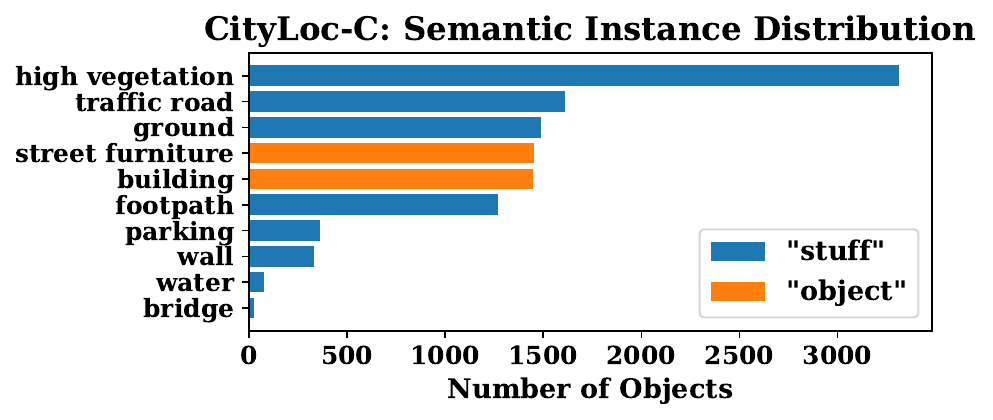}
      \caption{Semantic instance distribution in \textit{CityLoc-C}.}
  \label{fig:cityloc_c_distribution}
  \vspace{-1.5em}
\end{figure}

\para{Distribution of semantic instances.} For all semantic instances contained in the point cloud maps, we compute their distribution, as shown in~\cref{fig:cityloc_k_distribution} and~\cref{fig:cityloc_c_distribution}. Here, ``object'' and ``stuff'' follow the definitions in KITTI-360~\cite{liao2022kitti}, referring to countable and uncountable instances, respectively, and are rendered in orange and blue.

\section{Implementation Details}
\label{sec:implementation_details} 

\subsection{Textual Query Generation}
\textbf{Components.} \quad Each textual query $\mathcal{T}$ contains $N_t$ hints. 
Each hint $h$ describes an object by specifying its color, semantic category, and its directional relation with respect to the query location, following the template format: 
\texttt{"The pose is <direction> of <color> <semantic>."} 

\para{Direction computation.}
For each object $o$ referenced in the textual description, we first project all its 3D points onto the horizontal plane and identify the closest point $\mathbf{p}_{\text{close}}\in\mathbb{R}^2$ to the query position $\xi$.  If the distance between $\mathbf{p}_{\text{close}}$ and $\xi$ is below $\delta = 2.5$ m, the direction is assigned as ``on-top''. 
Otherwise, we determine the relative orientation by comparing the horizontal offsets $dx$ and $dy$ between $\xi$ and the centroid of $o$, as summarized in \cref{eq:direction}.

\begin{equation}
\text{Direction}(\xi,o)=
\begin{cases}
\text{``on-top''}, 
& \|(\xi-\mathbf{p}_{\text{close}})\|_2 < \delta, \\[5pt]
\text{``east''},  
& |dx| \ge |dy|\ \text{and}\ dx \ge 0, \\[3pt]
\text{``west''},  
& |dx| \ge |dy|\ \text{and}\ dx < 0, \\[3pt]
\text{``north''}, 
& |dx| < |dy|\ \text{and}\ dy \ge 0, \\[3pt]
\text{``south''}, 
& |dx| < |dy|\ \text{and}\ dy < 0,
\end{cases}
\label{eq:direction}
\end{equation}

\para{Color mapping.}
Each object in the map is already associated with an RGB color $\bar{\mathbf{c}}$ (as described in Eq.~(\ref{eq:color}) in the main paper). 
To obtain its textual color attribute, we map the object's RGB value to the nearest entry in a predefined discrete color palette 
$\texttt{COLOR\_NAMES}=\{\text{dark-green, gray, gray-green, bright-gray, black, green, beige}\}$ as in KITTI360Pose~\citep{kolmet2022text2pos}, 
using the corresponding template RGB centers $\mathcal{C}=\{\mathbf{c}_1,\dots,\mathbf{c}_K\}$ for nearest-neighbor assignment. The assigned textual label is obtained by
\begin{equation}
\label{eq:color_selection}
\text{color}(o)=\texttt{COLOR\_NAMES}\!\left[\arg\min_{k}\|\bar{\mathbf{c}}-\mathbf{c}_k\|\right].
\end{equation}
where $k \in \{1,\dots,K\}$.

Constructing large-scale, free-form textual descriptions from point cloud data is highly challenging and costly. 
Due to the complexity of real-world point cloud scenes, manually extracting scene information to compose detailed descriptions is prohibitively expensive. 
Although some recent methods~\cite{chu2024towards,ye2025cross} attempt to generate text by first extracting keywords and then prompting a large language model (LLM) (\eg, GPT-4~\cite{achiam2023gpt}). This process is generally not free and still requires manual verification, making it unsuitable for scalable dataset construction.

For these reasons, we follow prior work~\cite{kolmet2022text2pos} and adopt a rule-based template to encode the essential object-level cues for localization, including color, semantic category, and relative direction. This approach allows us to construct informative and scalable textual descriptions at minimal cost. It is fully free to generate, requires no LLMs, and avoids any manual post-verification, while still preserving the key information necessary for fine-grained localization.

\begin{table*}[!t]
    \centering
    \caption{System prompt for VLM-Loc.}
    \vspace{-0.3cm}
    \renewcommand{\arraystretch}{1.1}
    \tcbset{
        colback=gray!10,
        colframe=gray!50,
        arc=1mm,
        boxrule=0.25mm,
        width=\textwidth,
        left=3pt,
        right=3pt,
        top=2pt,
        bottom=2pt
    }
    \begin{tcolorbox}
        \textbf{You are a helpful assistant for spatial reasoning and grounding in BEV (bird's-eye-view) images.}
        You are also given a scene graph describing the environment as:
        \{node\_id, label, pixel\_center\}, where:
        - label = semantic label (e.g., ``road'', ``vegetation'', ``building'', ``pole'', etc.)
        - pixel\_center = pixel coordinates [x, y] (origin at top-left, x→right, y→down). \\[4pt]

        \textbf{COORDINATE \& DIRECTION RULES:}
        In BEV pixel coordinates:
        Up = North = y decreases;
        Down = South = y increases;
        Left = West = x decreases;
        Right = East = x increases. \\[4pt]

        \textbf{GOAL:}
        (1) Parse the user's natural-language description of a target location.
        (2) Extract all mentioned object phrases (e.g., “terrain'', “parking'') in order of appearance.
        (3) For each object phrase:
        Find the matching node in the given scene graph,
        determine whether grounding is successful (boolean),
        and record the matched node’s ID if grounded; otherwise set it to None.
        (4) Finally, infer the 2D pixel coordinate of the \textbf{target location}
        based on the described spatial relations (north/south/east/west/on-top). \\[4pt]

        \textbf{RULES:}
        All reasoning and distance computation MUST be in pixel coordinates.
        Length of assignments MUST equal the number of mentioned objects in the user's text.
        Each matched node\_id must exist in the scene graph;
        if not found or out of context, set {"grounded": false} and {"matched\_node": None}.
        Output exactly ONE JSON object in the schema below — \textbf{no explanations, no extra text}. \\[4pt]

        \textbf{OUTPUT FORMAT (STRICT):}
        \{
        "assignments": [
        {"object\_label": \textless string\textgreater,
        "grounded": \textless bool\textgreater,
        "matched\_node": \textless int\textbar None\textgreater}, ...],
        "point\_2d": [\textless int\textgreater, \textless int\textgreater]
        \} \\[4pt]

        \textbf{EXAMPLE (ILLUSTRATION ONLY):}
        OUTPUT:
        \{
        "assignments": [
        {"object\_label": "parking", "grounded": true, "matched\_node": 0},
        {"object\_label": "terrain", "grounded": true, "matched\_node": 8},
        {"object\_label": "road", "grounded": true, "matched\_node": 4},
        {"object\_label": "vegetation", "grounded": true, "matched\_node": 11}],
        "point\_2d": [45, 135]
        \}
    \end{tcolorbox}
    \label{tab:system_prompt}
\end{table*}

\subsection{System Prompt}

Our system prompt consists of six components designed to guide the VLM in performing the T2P localization task, as detailed in Tab.~\ref{tab:system_prompt}. The prompt could be formulated as :

\begin{itemize}
    \item \textbf{Role:} specifies the localization task for the VLM and clearly defines the expected input structure, including the types of information provided, their format, and the order in which they are supplied.
    \item \textbf{Coordinate system}: provides the correspondence between the pixel coordinate system and the geospatial coordinate system, which is essential for enabling the model to interpret directional cues in the text and map them to the correct orientation on the map.
    \item \textbf{Goal:} guides the model in estimating the current pose step-by-step. The procedure is: 1) comprehend the input text description; 2) extract the textual cues by sequentially processing each mentioned object; 3) assess the visibility of described elements and form text-node matching pairs for valid entries; 4) estimate the target location based on the aggregated geographic information.
    \item \textbf{Rules:} specifies the constraints for the model's output, \eg, that distances should be measured in pixel coordinates. These rules ensure that the pose estimation adheres to a parsable format and avoids invalid results.
    \item \textbf{Output format:} defines the required output structure, mandating that the model presents its results for node assignment and pose estimation in a predefined format.
    \item \textbf{Example:} provides a concrete example that illustrates the predefined output format for the model to follow. Specifically, the output should first present the visibility and assignment status for each textual hint, followed by the estimated 2D pixel coordinates.
\end{itemize}

\section{Additional Experiments}
\label{sec:supp_exp}

\subsection{Results on KITTI360Pose}
To enable a fair comparison on KITTI360Pose~\cite{kolmet2022text2pos}, we follow its retrieval-then-localization protocol and adopt the same Top-1 retrieval results from CMMLoc~\cite{xu2025cmmloc} for localization. We retrain VLM-Loc for localization on KITTI360Pose and evaluate it on the test set. Other methods perform localization using their pretrained models (Text2Pos~\cite{kolmet2022text2pos} is not available). Results are reported in~\cref{tab:kitti360pose}. Under same retrieval protocol, VLM-Loc achieves competitive localization performance compared to CMMLoc.

\begin{table}[t]
\centering
\small
\setlength{\tabcolsep}{1.2pt}
\begin{tabular}{lccc | lccc}
\toprule
Method & R@5 & R@10 & R@15 & Method & R@5 & R@10 & R@15 \\
\midrule
Text2Pos & / & / & / & MNCL   & 33.71 & 50.33 & 54.69 \\
Text2Loc & 37.27 & 51.32 & 54.79 & CMMLoc & \underline{37.81} & \textbf{51.84} & \textbf{55.02} \\
\midrule
\rowcolor{gray!20}
\multicolumn{1}{l}{\textbf{VLM-Loc}} & \textbf{40.36} & \underline{51.69} & \underline{54.74} \\
\bottomrule
\end{tabular}
\vspace{-2.5mm}
\caption{Localization on KITTI360Pose test set (11404 samples).}
\label{tab:kitti360pose}
\end{table}

\subsection{Inference Analysis}
Inference analysis is conducted on two RTX 4090 GPUs with the batch size of 1, and results are reported in \cref{tab:inference_analysis}. The latency is acceptable for the T2P localization setting. Moreover, the inference cost can be significantly reduced via quantization, smaller backbones, and optimized deployment frameworks, making VLM-Loc more practical for real-world systems.

\begin{table}[!t]
\centering
\small
\setlength{\tabcolsep}{3pt}
\begin{tabular}{l | ccc}
\toprule
Backbone & FPS & Peak Mem. (GB) & Params (B) \\
\midrule
Qwen-VL-2B-Instruct & 0.36 & 8.50 & 2.14 \\
Qwen-VL-8B-Instruct & 0.23 & 33.65 & 8.79 \\
\bottomrule
\end{tabular}
\caption{
Inference analysis of VLM-Loc on the \textit{CityLoc-K} val set.
}
\label{tab:inference_analysis}
\end{table}

\section{Qualitative Results}
\label{sec:more_vis} 
    
\begin{figure*}[!t]
  \centering
  \includegraphics[width=0.95\textwidth]{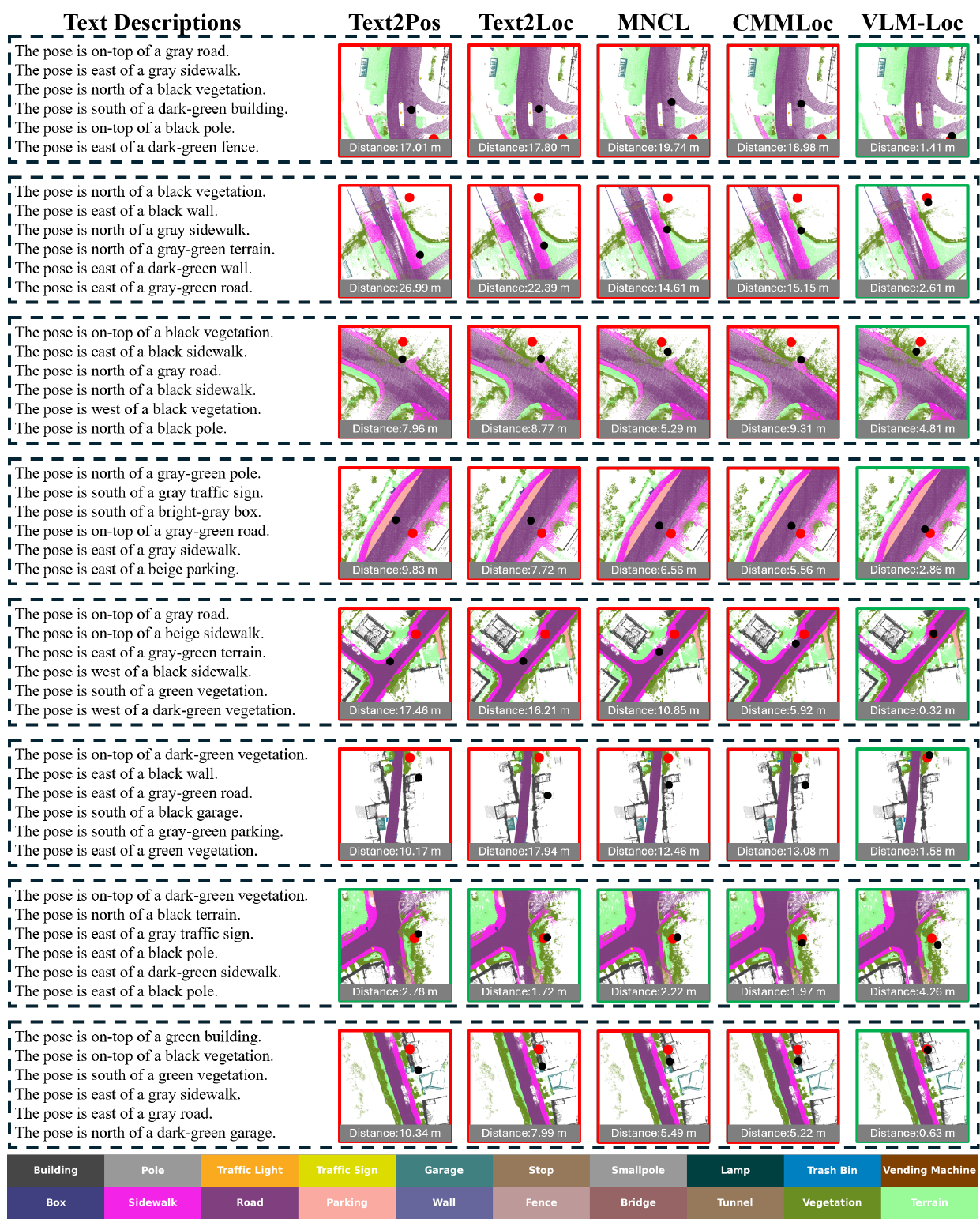}
  \caption{Qualitative results of VLM-Loc and baseline methods on the \textit{CityLoc-K}.
Each example visualizes the predicted and GT positions on colorized BEV maps rendered with semantic labels.
The red circles \textcolor{red}{\ding{108}} and black circles \textcolor{black}{\ding{108}} denote the GT and predicted positions, respectively. The localization error is shown below each image, and green/red borders indicate localization error below/above 5 m.}
  \label{fig:appendix_vis_k}
\end{figure*}

\begin{figure*}[!t]
  \centering
  \includegraphics[width=0.95\textwidth]{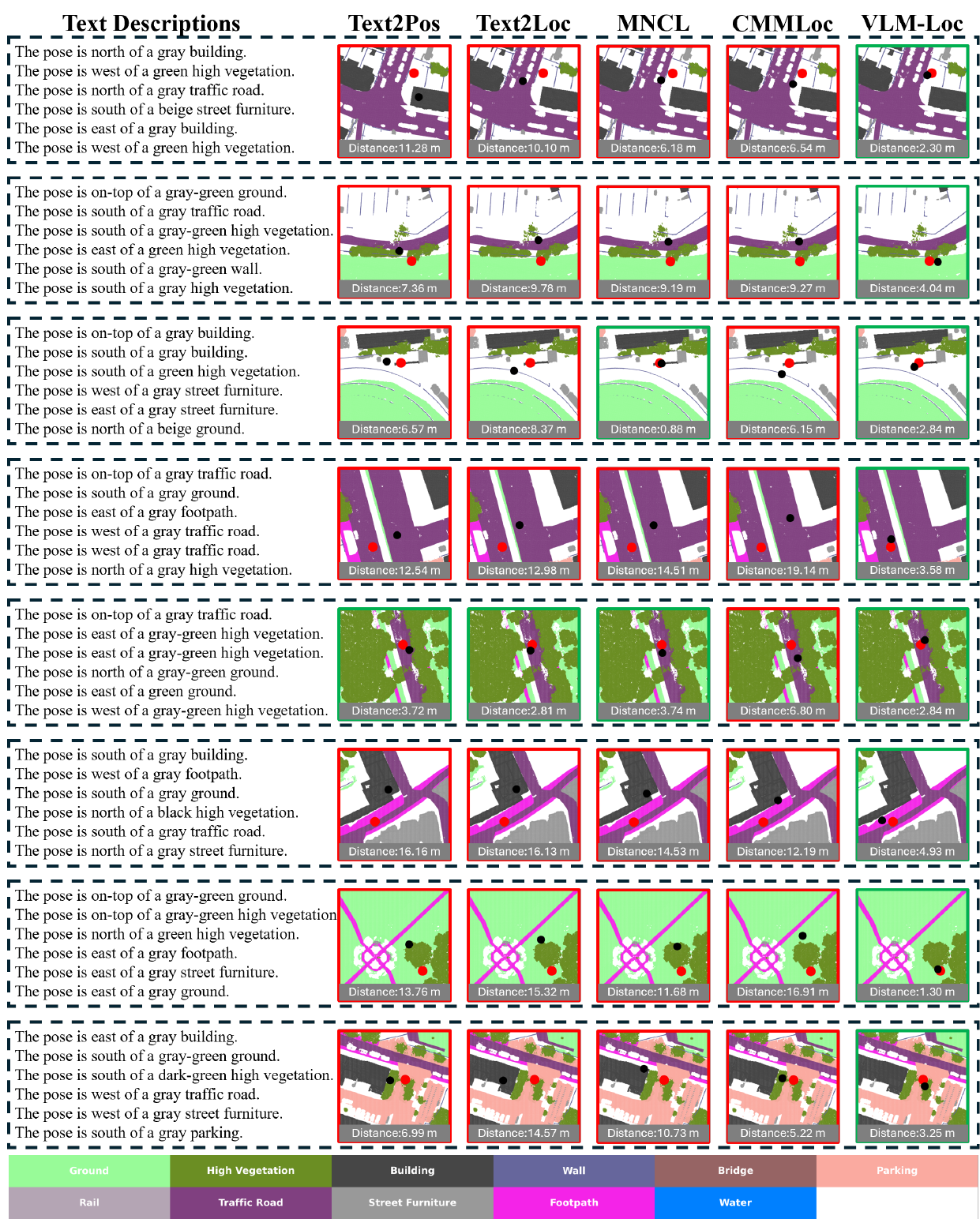}
  \caption{Qualitative results of VLM-Loc and baseline methods on the \textit{CityLoc-C}. Each example visualizes the predicted and GT positions on colorized BEV maps rendered with semantic labels.
The red circles \textcolor{red}{\ding{108}} and black circles \textcolor{black}{\ding{108}} denote the GT and predicted positions, respectively. The localization error is shown below each image, and green/red borders indicate localization error below/above 5 m.}
  \label{fig:appendix_vis_c}
\end{figure*}

Additional qualitative results of our approach and baselines on the \textit{CityLoc-K} and \textit{CityLoc-C} splits are shown in Fig.~\ref{fig:appendix_vis_k} and Fig.~\ref{fig:appendix_vis_c}. The results demonstrate the superior performance of the proposed method over baselines across diverse scenes, validating its robustness and accuracy.

\end{document}